\documentclass[letterpaper, 10 pt, conference]{ieeeconf} \usepackage{booktabs}
\usepackage{multirow}
\usepackage{graphicx}
\usepackage{stfloats}
\usepackage{amsmath}
\usepackage{mathtools}
\usepackage{float}
 \usepackage{amssymb}
 \usepackage{graphicx}   
\usepackage{multirow}
\let\labelindent\relax
\usepackage{enumitem}
\usepackage{cuted}
\usepackage{caption}
\IEEEoverridecommandlockouts                             
\title{\LARGE \bf
GroundControl: Anticipating Navigation Failures in Vision-Language Agents via Trajectory-Consistent Uncertainty Estimates}

\author{Nastaran Darabi$^{1}$, Divake Kumar$^{1}$, Sina Tayebati$^{1}$, Devashri Naik$^{1}$ and Amit Ranjan Trivedi $^{1}$ 
\thanks{ This work was supported in part by COGNISENSE, one of seven centers in JUMP 2.0, a Semiconductor Research Corporation (SRC) program sponsored by DARPA, and NSF funding \#2235207. Corresponding Authors Email:
ndarab2@uic.edu, amitrt@uic.edu}
\thanks{$^{1}$Authors are with the University of Illinois at Chicago (UIC).}%
}

\begin{document}
\maketitle
\thispagestyle{empty}
\pagestyle{empty}

\begin{strip}
    \centering
    \includegraphics[width=\linewidth]{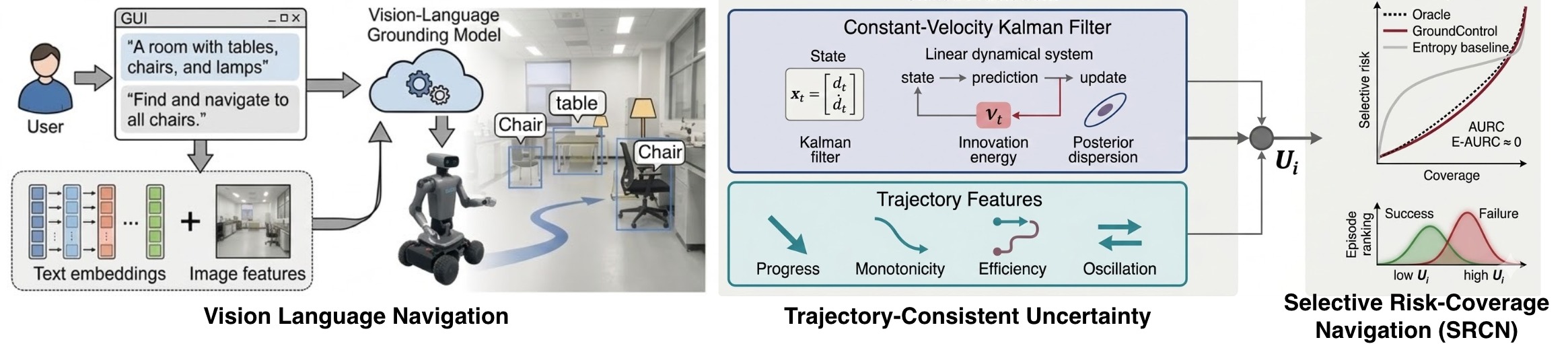}
    \captionof{figure}{
Overview of the \textbf{GroundControl} framework for trajectory-consistent uncertainty estimation in vision-language navigation. 
A VLM-based navigation agent produces a trajectory with distance-to-goal signal $\{d_t\}$ while executing actions conditioned on observations and language instructions. 
GroundControl models the evolution of this signal using a constant-velocity Kalman filter that predicts nominal goal-directed progress. 
Deviations from this prediction are captured through normalized innovation statistics and posterior covariance growth, and fused with trajectory descriptors including normalized progress, monotonicity, path efficiency, and oscillatory behavior. 
The resulting episode-level score $U_i$ measures geometric and temporal inconsistency in the trajectory rather than instantaneous decision ambiguity. 
Uncertainty quality is evaluated using the proposed Selective Risk--Coverage Navigation (SRCN) protocol, which assesses how effectively uncertainty ranks navigation episodes by failure or inefficiency using risk--coverage curves and AURC / E-AURC metrics.
}
    
    \label{overview}
\end{strip}

\begin{abstract}
Vision-language navigation agents achieve competitive average success on benchmark tasks, yet failures often arise through predictable trajectory-level breakdowns such as oscillation, stagnation, or inefficient detours. Reliable deployment, therefore, requires uncertainty signals that anticipate emerging failure dynamics during execution rather than reflect only instantaneous action entropy. We introduce \emph{GroundControl}, a trajectory-consistent uncertainty estimator defined as statistical deviation from nominal goal-directed distance-to-goal dynamics aggregated over an episode. GroundControl models distance evolution using a constant-velocity Kalman filter and combines normalized innovation statistics with complementary trajectory features capturing progress, monotonicity, path efficiency, and oscillatory behavior. The resulting uncertainty score reflects geometric and temporal inconsistency in navigation behavior rather than local prediction dispersion. To evaluate uncertainty quality independently of task success, we formalize \emph{Selective Risk--Coverage Navigation (SRCN)}, a protocol that measures how effectively an uncertainty score ranks episodes by failure or inefficiency using risk--coverage curves and AURC / E-AURC summaries. Across five EB-Navigation splits ($N=300$ episodes), trajectory-consistent uncertainty achieves near-oracle ordering under success-based selective risk, with weighted-average $\mathrm{E\text{-}AURC}_{\mathrm{SR}}=0.0024$ for the GPT-4o model, substantially outperforming entropy-, conformal-, and heuristic baselines. Under SPL-based selective evaluation, GroundControl consistently achieves the lowest AURC and E-AURC across models and navigation splits. These results show that modeling deviation from goal-directed dynamics provides an interpretable and robust signal for anticipating navigation failures in vision-language agents. 
\end{abstract}

\section{Introduction}
\label{sec:intro}

Vision-language navigation (VLN) agents have recently achieved strong benchmark performance by combining language understanding with visual grounding and discrete motion policies. Modern vision-language models (VLMs) can interpret multi-step instructions, resolve object references in egocentric observations, and execute navigation primitives with competitive Success weighted by Path Length (SPL) \cite{windecker2025navitrace, gao2024vision, liu2025embodied}. While these results indicate substantial progress in instruction following, benchmark success alone does not imply reliable execution. In practice, navigation failures often emerge gradually during execution through characteristic trajectory-level breakdowns. Agents may stall near distractors, oscillate between conflicting actions, accumulate path length without proportional geometric progress, or drift away from the intended goal. These behaviors are reflected in the temporal evolution of the distance-to-goal signal and action trajectory. However, current VLN systems rarely provide a principled mechanism for detecting such failure dynamics while an episode is unfolding \cite{zhao2023mind, zhu2020vision, wang2026vlingnav}.

For reliable deployment, navigation systems therefore require an uncertainty signal indicating whether an ongoing trajectory is deviating from successful goal-directed execution. Existing uncertainty proxies, however, are poorly suited for this setting. Most rely on instantaneous signals such as predictive entropy over action distributions or token-level confidence. These measures capture ambiguity in individual decisions but do not reflect whether the resulting trajectory remains consistent with geometric progress toward the goal. An agent may therefore maintain high step-wise confidence while repeatedly executing actions that lead to oscillation, stagnation, or inefficient detours.

This suggests that uncertainty in embodied navigation should reflect \textit{trajectory-level consistency of goal-directed dynamics}. In successful episodes, the distance-to-goal signal typically follows a structured evolution characterized by sustained progress with bounded variation. Systematic violations of this structure, such as oscillation, stagnation, divergence, or low path efficiency relative to displacement, provide quantitative evidence that execution is deviating from the intended navigation objective. Under this view, uncertainty estimation becomes the problem of detecting statistically significant deviations from expected goal-directed motion.

We implement this principle through \textbf{GroundControl}, an interpretable trajectory-consistent uncertainty estimator. GroundControl models distance-to-goal dynamics using a constant-velocity Kalman filter over the state $x_t = [d_t, \dot d_t]^\top$. The filter predicts nominal goal-directed progress and quantifies deviation through normalized innovation statistics and posterior covariance growth. These signals are fused with complementary trajectory features capturing normalized progress, monotonicity of the distance sequence, path efficiency relative to displacement, and action-reversal frequency. The resulting episode-level uncertainty score $U_i$ reflects temporal, geometric, and behavioral inconsistency in the trajectory rather than dispersion in individual action.

To evaluate uncertainty independently of raw task success, we introduce \textit{Selective Risk--Coverage Navigation (SRCN)}, a protocol for trajectory-level uncertainty signals that measures how effectively an uncertainty score ranks navigation episodes by failure using risk--coverage curves and summary metrics including AURC and excess-AURC. This formulation isolates ranking quality without modifying the underlying navigation policy and enables comparison across entropy and behavioral estimators. Our contributions are:

\begin{itemize}[leftmargin=*, itemsep=1pt, topsep=2pt]
\item We formalize trajectory-level consistency of distance-to-goal dynamics as a foundation for uncertainty estimation in VLN-based embodied navigation.

\item We introduce \textbf{GroundControl}, a lightweight trajectory-consistent estimator that detects statistically significant deviations from nominal goal-directed motion.

\item Across five EB-Navigation splits ($N=300$ episodes), our trajectory-consistent uncertainty achieves near-oracle ordering under success-based selective risk with weighted-average $\mathrm{E\text{-}AURC}_{\mathrm{SR}}=0.0024$, outperforming entropy-, conformal-, and heuristic baselines while remaining competitive under SPL-based selective evaluation.
\end{itemize}

\section{Relation to Prior Work}

\noindent\textbf{Embodied Navigation Benchmarks and Evaluation.}
Instruction-conditioned embodied navigation has been standardized through benchmarks such as Room-to-Room (R2R)~\cite{ku2020room} and Habitat-based environments~\cite{savva2019habitat, krantz2020beyond, raychaudhuri2024zero}. Agents follow natural language instructions using egocentric observations and are evaluated primarily by Success Rate and Success weighted by Path Length (SPL). Recent vision-language systems built on large pretrained transformers achieve competitive performance under these metrics \cite{zheng2025efficient, liu2025embodied}. However, they measure task outcome only after episode termination and do not assess whether an agent can anticipate failure during execution \cite{zhang2025embodied, hou2026survey, lin2025vlnverse}.

\vspace{2pt}
\noindent\textbf{Uncertainty in Sequential Decision Making.}
Uncertainty in embodied agents arises from partial observability, perceptual aliasing, ambiguous language grounding, and stochastic control. In VLN, common proxies include predictive entropy over action distributions, logit margins, and heuristics derived from reasoning traces. In reinforcement learning and robotics, epistemic and aleatoric uncertainty have been modeled using Bayesian critics, deep ensembles, and distributional value functions to support exploration or risk-sensitive control. These methods operate at the level of action probabilities or value estimates and are typically evaluated through state-wise prediction accuracy. None models distance-to-goal evolution as a stochastic dynamical process, and therefore they do not capture structured trajectory-level failure modes that unfold over multiple steps \cite{gawlikowski2023survey, yu2025uncertainty, tayebati2025learning}.

\vspace{2pt}
\noindent\textbf{Failure Prediction and Introspection.}
Failure prediction for embodied agents has been explored using auxiliary classifiers over latent representations, trajectory statistics such as episode length, and heuristic signals, including invalid-action rates or plan divergence. While these signals may correlate with failure on fixed benchmarks, they are often architecture-specific or implicitly exploit dataset artifacts, such as the coupling between failure and maximum episode length. In contrast, geometric observables such as distance-to-goal provide a task-level, model-agnostic signal of progress dynamics. Despite being readily available in navigation environments, trajectory-level geometric consistency has not been formalized for uncertainty estimation \cite{kondoh2025embodied, ahmed2021self, ye2021auxiliary}.

\vspace{2pt}
\noindent\textbf{Selective Prediction and Risk-Coverage Analysis.}
Selective prediction evaluates models that abstain on high-uncertainty inputs, trading coverage for reduced risk~\cite{tayebati2025learning}. Risk--coverage curves and metrics such as AURC or excess-AURC summarize ranking quality independently of threshold choice. Although widely adopted in classification, this framework is limited in embodied navigation, where uncertainty must be aggregated over temporally correlated trajectories rather than individual states. Our formulation adapts selective evaluation to episode-level uncertainty \cite{gawlikowski2023survey, anderson2018vision, savva2019habitat}.


\section{Methodology}
\label{sec:methodology}

\subsection{Task Setup and Standard Navigation Metrics}
\label{sec:task}

Each episode $i \in \{1,\dots,N\}$ specifies an initial robot pose, a natural-language instruction $\mathcal{L}_i$, and a target object definition~\cite{yang2025embodiedbench}. At each time step $t = 1,\dots,T_i$, the agent receives an egocentric observation $\mathbf{o}_t$ and selects a discrete action $\mathbf{u}_t \in \mathcal{A}$, where $|\mathcal{A}| = 8$ corresponds to forward, backward, left, right, rotate-left, rotate-right, look-up, and look-down motions of fixed magnitude.

An episode is considered successful if the agent reaches the target within a distance threshold $\delta = 1\,\text{m}$, denoted by $\mathrm{Succ}_i \in \{0,1\}$. In addition to Success Rate, we report Success weighted by Path Length (SPL),
\begin{equation}
  \mathrm{SPL}_i
  =
  \mathrm{Succ}_i \cdot
  \frac{p_i}{\max(p_i,\,l_i)},
  \label{eq:spl}
\end{equation}
where $p_i$ is the initial geodesic distance to the goal and $l_i = T_i \cdot \Delta$ is the executed path length with fixed step size $\Delta = 0.25\,\text{m}$. These metrics evaluate task completion and efficiency only at episode termination. In this work, we retain them unchanged and introduce complementary measures that assess the quality of episode-level uncertainty.

For each episode $i$, an uncertainty estimator produces a scalar score $U_i \in \mathbb{R}$, where lower values indicate higher confidence. The score may originate from internal state statistics such as posterior covariance or innovation energy, model-internal signals including attention entropy or belief dispersion, or post hoc behavioral measures such as action entropy, plan instability, invalid-action rates, or conformal nonconformity. 

This abstraction allows heterogeneous uncertainty estimators to be evaluated within a common framework while isolating the quality of their episode-level ranking. In particular, the SRCN evaluation introduced later depends only on the ordering induced by $U_i$ through thresholding.

\subsection{Selective Risk--Coverage Navigation (SRCN)}
\label{sec:srcn}

We formalize \emph{Selective Risk--Coverage Navigation} (SRCN) as an evaluation protocol for episode-level uncertainty ranking. Given uncertainty scores $\{U_i\}_{i=1}^N$, SRCN measures how well these scores order episodes by true loss without modifying the underlying navigation policy, where high-uncertainty episodes are progressively excluded, and performance is evaluated on the retained subset.

For a threshold $\tau$, define the retained index set
\begin{equation}
  \mathcal{I}(\tau) \triangleq \{\, i : U_i \le \tau \,\}, 
  \qquad
  \mathrm{Cov}(\tau) \triangleq \frac{|\mathcal{I}(\tau)|}{N},
  \label{eq:coverage}
\end{equation}
where $\mathrm{Cov}(\tau) \in [0,1]$ denotes coverage, i.e., the fraction of retained episodes. As $\tau$ increases, coverage increases monotonically to $1$.

\vspace{2pt}
\noindent\textbf{Selective risk.}
For success-based evaluation, we define the episode loss
$\ell_i^{\mathrm{SR}} = 1 - \mathrm{Succ}_i$. The selective risk at threshold $\tau$ is
\begin{equation}
  \mathrm{Risk}_{\mathrm{SR}}(\tau)
  \triangleq
  \frac{1}{|\mathcal{I}(\tau)|}
  \sum_{i \in \mathcal{I}(\tau)} \ell_i^{\mathrm{SR}},
  \label{eq:risk_sr}
\end{equation}
which equals the empirical failure rate among retained episodes. To incorporate path efficiency we also define the SPL-based loss
$\ell_i^{\mathrm{SPL}} = 1 - \mathrm{SPL}_i$, yielding
\begin{equation}
  \mathrm{Risk}_{\mathrm{SPL}}(\tau)
  \triangleq
  \frac{1}{|\mathcal{I}(\tau)|}
  \sum_{i \in \mathcal{I}(\tau)} \ell_i^{\mathrm{SPL}}.
  \label{eq:risk_spl}
\end{equation}
This variant penalizes both outright failures and inefficient successful trajectories.

\vspace{2pt}
\noindent\textbf{Risk--Coverage curve and AURC.}
Varying $\tau$ traces a risk--coverage curve that describes how risk decreases as increasingly uncertain episodes are removed. An uncertainty score with strong ranking quality yields low selective risk even at high coverage. We summarize this trade-off using the \emph{area under the risk--coverage curve} (AURC). Episodes are sorted in ascending order of $U_i$, and the integral
\begin{equation}
  \mathrm{AURC}_{\mathrm{SPL}}
  \triangleq
  \int_{0}^{1} \mathrm{Risk}_{\mathrm{SPL}}(\mathrm{Cov}) \, d\mathrm{Cov},
  \label{eq:aurc}
\end{equation}
is computed via the trapezoidal rule; $\mathrm{AURC}_{\mathrm{SR}}$ is defined analogously. Lower AURC indicates better ranking.

\vspace{2pt}
\noindent\textbf{Excess-AURC (E-AURC).}
Because AURC depends on task difficulty and class balance, we report the normalized \emph{excess AURC}
\begin{equation}
  \mathrm{E\text{-}AURC}
  \triangleq
  \mathrm{AURC} - \mathrm{AURC}^{\star},
  \label{eq:eaurc}
\end{equation}
where $\mathrm{AURC}^{\star}$ is the oracle lower bound obtained by sorting episodes by true loss. $\mathrm{E\text{-}AURC}=0$ corresponds to perfect ordering.

\noindent\textbf{Calibration.}
When a calibrated failure probability $\hat{p}_i(\mathrm{fail})$ is available, we additionally report Expected Calibration Error (ECE):
\begin{equation}
  \mathrm{ECE}
  \triangleq
  \sum_{m=1}^{M}
  \frac{|B_m|}{N}
  \bigl| \mathrm{acc}(B_m) - \mathrm{conf}(B_m) \bigr|,
  \label{eq:ece}
\end{equation}
where $\{B_m\}$ are uniform bins over $[0,1]$, $\mathrm{acc}(B_m)$ is the empirical failure frequency in bin $m$, and $\mathrm{conf}(B_m)$ is the average predicted failure probability.

\subsection{GroundControl: Uncertainty via Distance Dynamics}
\label{sec:groundcontrol}

GroundControl estimates trajectory-consistent uncertainty by modeling the evolution of distance-to-goal and measuring deviation from expected goal-directed dynamics. The key observation is that successful navigation exhibits locally consistent geometric progress, whereas oscillation, stagnation, or divergence produces statistically detectable inconsistency.

\vspace{2pt}
\noindent\textbf{Distance-to-Goal Dynamics via Constant-Velocity KF.}
Let $d_t$ denote the Euclidean distance from the agent to the goal at step $t$. Rather than modeling full spatial motion, we treat the scalar distance trajectory as a dynamical signal. Under nominal execution, distance evolves smoothly and decreases approximately linearly over short horizons. We capture this local structure using the state
\[
\mathbf{x}_t = [d_t, \dot{d}_t]^\top,
\]
where $\dot{d}_t$ denotes the instantaneous rate of change.

We adopt linear-Gaussian dynamics
\begin{equation}
  \mathbf{x}_{t+1} = \mathbf{F}\mathbf{x}_t + \mathbf{w}_t,
  \quad
  d_t = \mathbf{H}\mathbf{x}_t + v_t,
\end{equation}
with
\[
\mathbf{F} =
\begin{bmatrix}
1 & 1 \\
0 & 1
\end{bmatrix},
\quad
\mathbf{H} = [1 \; 0],
\]
process noise $\mathbf{w}_t \sim \mathcal{N}(0,\mathbf{Q})$ and observation noise $v_t \sim \mathcal{N}(0,\sigma_r^2)$. The constant-velocity assumption is not intended to model exact robot motion; it provides a locally consistent surrogate for nominal goal-directed progress against which deviations can be tested statistically.

The filter is initialized as
\[
\hat{\mathbf{x}}_0 = [d_0, 0]^\top,
\qquad
\boldsymbol{\Sigma}_0 = \mathrm{diag}(0.01, 1.0),
\]
encoding high confidence in initial distance and diffuse uncertainty in velocity.

At each step, standard Kalman prediction and update equations are applied:
\begin{align}
  \hat{\mathbf{x}}_{t|t-1} &= \mathbf{F}\hat{\mathbf{x}}_{t-1|t-1}, \\
  \boldsymbol{\Sigma}_{t|t-1} &= \mathbf{F}\boldsymbol{\Sigma}_{t-1|t-1}\mathbf{F}^\top + \mathbf{Q}, \\
  \nu_t &= d_t - \mathbf{H}\hat{\mathbf{x}}_{t|t-1}, \\
  S_t &= \mathbf{H}\boldsymbol{\Sigma}_{t|t-1}\mathbf{H}^\top + \sigma_r^2, \\
  \mathbf{K}_t &= \boldsymbol{\Sigma}_{t|t-1}\mathbf{H}^\top / S_t, \\
  \hat{\mathbf{x}}_{t|t} &= \hat{\mathbf{x}}_{t|t-1} + \mathbf{K}_t \nu_t, \\
  \boldsymbol{\Sigma}_{t|t} &=
    (\mathbf{I}-\mathbf{K}_t\mathbf{H})\boldsymbol{\Sigma}_{t|t-1}
    (\mathbf{I}-\mathbf{K}_t\mathbf{H})^\top
    + \sigma_r^2 \mathbf{K}_t\mathbf{K}_t^\top.
\end{align}

The innovation
\[
\nu_t = d_t - \mathbf{H}\hat{\mathbf{x}}_{t|t-1}
\]
measures the discrepancy between observed and predicted distance. The normalized innovation squared
\(
\frac{\nu_t^2}{S_t}
\)
is the squared Mahalanobis residual and, under linear-Gaussian assumptions, follows a $\chi^2$ distribution with one degree of freedom. It therefore serves as a statistical consistency test: small values indicate agreement with nominal goal-directed dynamics, whereas persistently large values indicate significant deviation from expected geometric progress.

\vspace{2pt}
\noindent\textbf{Per-Step Uncertainty Score.}
We define a per-step uncertainty score $\mathcal{H}_t$ that aggregates three complementary indicators of trajectory inconsistency: (i) posterior belief dispersion, (ii) action-level instability, and (iii) statistical deviation from nominal goal-directed dynamics:
\begin{equation}
  \mathcal{H}_t \triangleq
    \lambda_1 \log\det\!\bigl(\boldsymbol{\Sigma}_{t|t} + \varepsilon\mathbf{I}\bigr)
    + \lambda_2\,\mathcal{H}(\alpha_t)
    + \lambda_3\,\frac{\nu_t^2}{S_t}.
  \label{eq:Ht}
\end{equation}
The term $\log\det(\boldsymbol{\Sigma}_{t|t})$ measures posterior belief dispersion and increases when the filter becomes uncertain about local distance dynamics. The normalized innovation squared $\nu_t^2 / S_t$ is the squared Mahalanobis residual; under linear-Gaussian consistency assumptions, it follows a $\chi^2$ distribution with one degree of freedom and therefore acts as a statistical consistency test. Persistently large values indicate significant deviation from expected goal-directed motion.

The action-level term $\mathcal{H}(\alpha_t)$ is the Bernoulli entropy
\begin{equation}
    \mathcal{H}(\alpha_t) = -\alpha_t\log\alpha_t - (1{-}\alpha_t)\log(1{-}\alpha_t),
\end{equation}
where $\alpha_t \in [0,1]$ is the normalized confidence assigned by the policy to the executed action at time $t$ (i.e., the softmax probability of the selected action before environment execution). The entropy is maximal near $0.5$ and minimal near $0$ or $1$, capturing action-level indecision. The weights $(\lambda_1, \lambda_2, \lambda_3) = (1.0,\,0.5,\,0.8)$ are fixed across all experiments to isolate signal quality from learned weighting effects.

\vspace{5pt}
\noindent\textbf{Multi-Signal Episode Uncertainty.}
To obtain an episode-level signal, we aggregate the per-step score by taking the maximum over time and mapping it to $[0,1]$ via a sigmoid:
\begin{equation}
\tilde{s}^{\mathrm{skf}}_i = \sigma\!\left(\max_t \mathcal{H}_t\right).
\end{equation}
The max operator preserves brief but statistically significant deviations that might otherwise be diluted by stable segments, while the sigmoid normalizes scale across episodes. Thus $\tilde{s}^{\mathrm{skf}}_i$ reflects peak dynamical inconsistency within episode $i$.

We fuse $\tilde{s}^{\mathrm{skf}}_i$ with four complementary trajectory-level features, each normalized to $[0,1]$ with larger values indicating greater deviation from nominal goal-directed motion:
\begin{align}
  f_i^{\mathrm{prog}} &= 1 - \mathrm{clip}\!\left(
    \frac{d_0^{(i)} - d_{T_i}^{(i)}}{d_0^{(i)}},\,0,\,1\right),
  \label{eq:fprog}\\[4pt]
  f_i^{\mathrm{mono}} &= 1 - \frac{1}{T_i-1}
    \sum_{t=1}^{T_i-1}\mathbb{1}[d_t < d_{t-1}],
  \label{eq:fmono}\\[4pt]
  f_i^{\mathrm{eff}} &= 1 - \frac{|d_0^{(i)} - d_{T_i}^{(i)}|}
    {\sum_{t=0}^{T_i-1}|d_{t+1} - d_t|},
  \label{eq:feff}\\[4pt]
  f_i^{\mathrm{osc}} &= \frac{1}{T_i-1}
    \sum_{t=1}^{T_i-1}\mathbb{1}[(\mathbf{u}_{t-1},\mathbf{u}_t)\in
    \mathcal{R}],
  \label{eq:fosc}
\end{align}
where $\mathcal{R}$ denotes action-reversal pairs. These features capture insufficient net progress, non-monotonic distance evolution, geometric inefficiency relative to cumulative motion, and oscillatory control behavior.

The final episode-level uncertainty score is defined as a fixed linear fusion
\begin{equation}\small
  U_i \triangleq
    w_{\mathrm{skf}}\,\tilde{s}^{\mathrm{skf}}_i
    + w_{\mathrm{prog}}\,f_i^{\mathrm{prog}}
    + w_{\mathrm{mono}}\,f_i^{\mathrm{mono}}
    + w_{\mathrm{eff}}\,f_i^{\mathrm{eff}}
    + w_{\mathrm{osc}}\,f_i^{\mathrm{osc}},
  \label{eq:Ui}
\end{equation}
with weights $(w_{\mathrm{skf}}, w_{\mathrm{prog}}, w_{\mathrm{mono}}, w_{\mathrm{eff}}, w_{\mathrm{osc}}) = (0.30, 0.25, 0.20, 0.15, 0.10)$. Fixed weights preserve interpretability and isolate the intrinsic quality of trajectory-consistent signals from learned weighting effects.


\begin{table}[t]
\centering
\caption{Standard navigation performance on EB-Navigation.\vspace{-5pt}}
\label{tab:nav_perf_combined}
\resizebox{\linewidth}{!}{
\begin{tabular}{lcc|cc|cc}
\toprule
& \multicolumn{2}{c|}{GPT-4o} & \multicolumn{2}{c|}{GPT-5-mini} & \multicolumn{2}{c}{Gemini-1.5-Flash} \\
Set & Success (\%) & SPL & Success (\%) & SPL & Success (\%) & SPL \\
\midrule
base & 58.3 & 0.493 & 75.0 & 0.656 & 60.0 & 0.538 \\
common\_sense & 58.3 & 0.497 & 70.0 & 0.618 & 55.0 & 0.475 \\
complex\_inst. & 56.7 & 0.509 & 73.3 & 0.668 & 46.7 & 0.408 \\
visual\_app. & 53.3 & 0.464 & 65.0 & 0.576 & 53.7 & 0.446 \\
long\_horizon & 16.7 & 0.112 & 16.7 & 0.129 & 14.5 & 0.107 \\
\bottomrule\vspace{-25pt}
\end{tabular}}
\end{table}


\begin{table*}[t]
\caption{Success-based selective risk evaluation (SRCN) across EB-Navigation splits. Metrics report $\mathrm{AURC}_{\mathrm{SR}}$ and $\mathrm{E\text{-}AURC}_{\mathrm{SR}}$ (lower is better). $\mathrm{E\text{-}AURC}=0$ corresponds to oracle ordering. \textbf{Best} results are highlighted in bold.\vspace{-5pt}}
\label{tab:srcn_sr}
\centering
\resizebox{\linewidth}{!}{
\renewcommand{\arraystretch}{1.05}

\begin{tabular}{c l|cccccc|cccccc|ccccc}
\toprule
& & \multicolumn{6}{c|}{GPT-4o} & \multicolumn{6}{c|}{GPT-5-mini} & \multicolumn{5}{c}{Gemini-1.5-Flash} \\
Metric & Method
& base & common & complex & visual & long & Avg
& base & common & complex & visual & long & Avg
& base & common & complex & visual & Avg \\
\midrule

\multirow{7}{*}{\rotatebox{90}{$\mathrm{AURC}_{\mathrm{SR}}\downarrow$}}

& GroundControl
& \textbf{0.10} & \textbf{0.10} & \textbf{0.11} & \textbf{0.13} & \textbf{0.55} & \textbf{0.20}
& \textbf{0.03} & \textbf{0.05} & \textbf{0.04} & \textbf{0.07} & \textbf{0.54} & \textbf{0.15}
& \textbf{0.09} & \textbf{0.12} & \textbf{0.18} & \textbf{0.13} & \textbf{0.13} \\

& Conformal
& 0.53 & 0.59 & 0.50 & 0.54 & 0.83 & 0.60
& 0.34 & 0.38 & 0.45 & 0.40 & 0.78 & 0.47
& 0.46 & 0.53 & 0.54 & 0.35 & 0.48 \\

& Entropy
& 0.25 & 0.21 & 0.22 & 0.34 & 0.90 & 0.38
& 0.04 & 0.06 & \textbf{0.04} & 0.08 & 0.65 & 0.17
& 0.27 & 0.22 & 0.28 & 0.23 & 0.25 \\


& Self-Consist.
& 0.30 & 0.21 & 0.21 & 0.29 & 0.86 & 0.37
& 0.07 & 0.09 & 0.08 & 0.10 & 0.80 & 0.23
& 0.18 & 0.27 & 0.30 & 0.24 & 0.25 \\

& Invalid-act
& 0.24 & 0.29 & 0.27 & 0.33 & 0.82 & 0.39
& 0.08 & 0.09 & 0.10 & 0.11 & 0.73 & 0.22
& 0.18 & 0.21 & 0.27 & 0.26 & 0.23 \\

& Random
& 0.33 & 0.43 & 0.38 & 0.42 & 0.83 & 0.48
& 0.18 & 0.22 & 0.23 & 0.34 & 0.79 & 0.35
& 0.32 & 0.40 & 0.44 & 0.42 & 0.39 \\

\midrule

\multirow{7}{*}{\rotatebox{90}{$\mathrm{E\text{-}AURC}_{\mathrm{SR}}\downarrow$}}

& GroundControl
& \textbf{0.00} & \textbf{0.00} & \textbf{0.00} & \textbf{0.00} & \textbf{0.01} & \textbf{0.00}
& \textbf{0.00} & \textbf{0.00} & \textbf{0.00} & \textbf{0.00} & \textbf{0.01} & \textbf{0.00}
& \textbf{0.00} & \textbf{0.00} & \textbf{0.00} & \textbf{0.00} & \textbf{0.00} \\

& Conformal
& 0.43 & 0.49 & 0.39 & 0.41 & 0.29 & 0.40
& 0.31 & 0.33 & 0.41 & 0.33 & 0.25 & 0.33
& 0.37 & 0.41 & 0.37 & 0.22 & 0.35 \\

& Entropy
& 0.15 & 0.11 & 0.11 & 0.21 & 0.36 & 0.19
& \textbf{0.00} & 0.01 & \textbf{0.00} & 0.01 & 0.12 & 0.03
& 0.17 & 0.10 & 0.10 & 0.10 & 0.12 \\


& Self-Consist.
& 0.20 & 0.11 & 0.10 & 0.16 & 0.33 & 0.18
& 0.04 & 0.04 & 0.04 & 0.03 & 0.26 & 0.08
& 0.09 & 0.15 & 0.13 & 0.11 & 0.12 \\

& Invalid-act
& 0.13 & 0.19 & 0.15 & 0.20 & 0.29 & 0.19
& 0.05 & 0.04 & 0.06 & 0.04 & 0.20 & 0.08
& 0.09 & 0.09 & 0.09 & 0.13 & 0.10 \\

& Random
& 0.23 & 0.33 & 0.27 & 0.28 & 0.29 & 0.28
& 0.15 & 0.17 & 0.19 & 0.27 & 0.26 & 0.21
& 0.23 & 0.28 & 0.26 & 0.29 & 0.26 \\

\bottomrule
\end{tabular}}
\end{table*}

\begin{table*}[t]
\caption{SPL-based selective risk evaluation (SRCN) across EB-Navigation splits. Metrics report $\mathrm{AURC}_{\mathrm{SPL}}$ and $\mathrm{E\text{-}AURC}_{\mathrm{SPL}}$ (lower is better). $\mathrm{E\text{-}AURC}=0$ corresponds to oracle ordering. \textbf{Best} results are highlighted in bold. \vspace{-5pt}}
\label{tab:srcn_spl_combined}
\centering
\resizebox{\linewidth}{!}{
\renewcommand{\arraystretch}{1.05}

\begin{tabular}{c l|cccccc|cccccc|ccccc}
\toprule
& & \multicolumn{6}{c|}{GPT-4o} & \multicolumn{6}{c|}{GPT-5-mini} & \multicolumn{5}{c}{Gemini-1.5-Flash} \\
Metric & Method
& base & common & complex & visual & long & Avg
& base & common & complex & visual & long & Avg
& base & common & complex & visual & Avg \\
\midrule

\multirow{7}{*}{\rotatebox{90}{$\mathrm{AURC}_{\mathrm{SPL}}\downarrow$}}

& GroundControl & \textbf{0.19} & \textbf{0.18} & \textbf{0.17} & \textbf{0.21} & \textbf{0.67} & \textbf{0.29}
                & 0.12 & 0.12 & \textbf{0.09} & \textbf{0.13} & \textbf{0.63} & \textbf{0.22}
                & \textbf{0.16} & \textbf{0.20} & \textbf{0.26} & \textbf{0.23} & \textbf{0.21} \\

& Conformal & 0.61 & 0.64 & 0.55 & 0.61 & 0.88 & 0.66
             & 0.43 & 0.45 & 0.51 & 0.46 & 0.83 & 0.53
             & 0.53 & 0.59 & 0.59 & 0.45 & 0.55 \\

& Entropy & 0.32 & 0.27 & 0.26 & 0.39 & 0.92 & 0.43
           & \textbf{0.10} & \textbf{0.11} & \textbf{0.09} & 0.14 & 0.71 & 0.23
           & 0.31 & 0.27 & 0.33 & 0.31 & 0.30 \\


& Self-Consist. & 0.36 & 0.30 & 0.27 & 0.37 & 0.90 & 0.44
                & 0.16 & 0.17 & 0.14 & 0.17 & 0.83 & 0.30
                & 0.26 & 0.34 & 0.36 & 0.32 & 0.32 \\

& Invalid-act & 0.30 & 0.35 & 0.30 & 0.39 & 0.87 & 0.44
               & 0.16 & 0.15 & 0.15 & 0.17 & 0.77 & 0.28
               & 0.24 & 0.29 & 0.34 & 0.34 & 0.30 \\

& Random & 0.43 & 0.49 & 0.42 & 0.47 & 0.87 & 0.54
         & 0.27 & 0.29 & 0.29 & 0.41 & 0.83 & 0.42
         & 0.38 & 0.47 & 0.50 & 0.55 & 0.47 \\

\midrule

\multirow{7}{*}{\rotatebox{90}{$\mathrm{E\text{-}AURC}_{\mathrm{SPL}}\downarrow$}}

& GroundControl & \textbf{0.01} & \textbf{0.01} & \textbf{0.01} & \textbf{0.01} & \textbf{0.01} & \textbf{0.01}
                & 0.03 & 0.02 & 0.02 & \textbf{0.01} & \textbf{0.01} & \textbf{0.02}
                & \textbf{0.01} & \textbf{0.02} & \textbf{0.02} & \textbf{0.01} & \textbf{0.02} \\

& Conformal & 0.43 & 0.47 & 0.39 & 0.41 & 0.22 & 0.38
             & 0.34 & 0.35 & 0.43 & 0.34 & 0.22 & 0.34
             & 0.39 & 0.41 & 0.35 & 0.24 & 0.36 \\

& Entropy & 0.14 & 0.10 & 0.11 & 0.19 & 0.26 & 0.16
           & \textbf{0.01} & \textbf{0.01} & \textbf{0.01} & 0.02 & 0.09 & 0.03
           & 0.17 & 0.09 & 0.09 & 0.10 & 0.11 \\


& Self-Consist. & 0.18 & 0.13 & 0.12 & 0.17 & 0.24 & 0.17
                & 0.07 & 0.07 & 0.06 & 0.05 & 0.22 & 0.10
                & 0.11 & 0.15 & 0.12 & 0.11 & 0.13 \\

& Invalid-act & 0.13 & 0.18 & 0.15 & 0.19 & 0.21 & 0.17
               & 0.07 & 0.05 & 0.08 & 0.05 & 0.15 & 0.08
               & 0.10 & 0.10 & 0.11 & 0.13 & 0.11 \\

& Random & 0.25 & 0.32 & 0.26 & 0.28 & 0.21 & 0.26
         & 0.18 & 0.19 & 0.21 & 0.29 & 0.21 & 0.22
         & 0.23 & 0.28 & 0.26 & 0.34 & 0.28 \\

\bottomrule
\end{tabular}}
\end{table*}

\vspace{5pt}
\noindent\textbf{Interpretation.}
The four trajectory features capture complementary failure modes. 
$f^{\mathrm{prog}}$ identifies insufficient net reduction in distance-to-goal, 
$f^{\mathrm{mono}}$ penalizes frequent increases in distance indicating inconsistent approach dynamics, 
$f^{\mathrm{eff}}$ detects inefficient detours relative to displacement, and 
$f^{\mathrm{osc}}$ captures repeated action reversals indicative of unstable control. These features complement the Kalman-based statistical signals. Innovation energy detects local deviations from nominal distance dynamics, while posterior dispersion reflects uncertainty in distance evolution. Together, they capture geometric, temporal, and behavioral inconsistencies that single-signal baselines cannot represent.

\vspace{2pt}
\noindent\textbf{Active Recovery.}
Although our primary focus is uncertainty evaluation, the trajectory-consistent signal can also support online control. Let $\hat{U}_t$ denote the running episode score defined as the maximum per-step signal observed up to time $t$. A simple switching controller is
\begin{equation}
  \mathbf{u}_t =
  \begin{cases}
    \pi_{\mathrm{task}}(\mathbf{o}_{1:t}, \mathcal{L})
      & \hat{U}_t < \tau_{\mathrm{stable}},\\
    \pi_{\mathrm{recover}}(\mathbf{o}_{1:t}, \mathcal{L})
      & \tau_{\mathrm{stable}} \le \hat{U}_t < \tau_{\mathrm{crit}},\\
    \pi_{\mathrm{halt}}
      & \hat{U}_t \ge \tau_{\mathrm{crit}}.
  \end{cases}
  \label{eq:switching}
\end{equation}
where $(\tau_{\mathrm{stable}}, \tau_{\mathrm{crit}}) = (0.35, 0.65)$. Thus, trajectory-consistent uncertainty can also enable runtime introspection, though recovery policies are not studied in this paper.

\subsection{Baselines}
\label{sec:baselines}

We compare against seven representative uncertainty baselines spanning conformal, entropy-based, trajectory-based, and heuristic signals. Each baseline produces an episode-level score $U_i$ evaluated under the SRCN protocol.

\vspace{2pt}
\noindent\textbf{Conformal (ACI).}
Online adaptive conformal inference~\cite{gibbs2021adaptive} is implemented with step-wise nonconformity $|\hat{d}_t - d_t|$, where $\hat{d}_t = 2d_{t-1} - d_{t-2}$ extrapolates the distance trajectory. The episode score combines running miscoverage (60\%) and mean normalized residual (40\%).

\vspace{2pt}
\noindent\textbf{Predictive Entropy.}
Normalized Shannon entropy of the episode action histogram, $H(\mathbf{p})/\log|\mathcal{A}|$, measuring dispersion in action usage.


\vspace{2pt}
\noindent\textbf{Self-Consistency.}
Plan instability is measured as $1-\overline{J}$, where $\overline{J}$ is the mean Jaccard similarity between consecutive executable plans extracted from VLM reasoning~\cite{wang2022self}.


\vspace{2pt}
\noindent\textbf{Invalid-action rate.}
Fraction of steps where the executed action is rejected by the environment.

\vspace{2pt}
\noindent\textbf{Random.}
Random uncertainty scores $U_i \sim \mathrm{Uniform}(0,1)$ as a lower bound.

\section{RESULTS}
\label{sec:results}

\subsection{Experimental Protocol and Sets}

We evaluate episode-level uncertainty quality on EB-Navigation across five splits:
\texttt{base}, \texttt{common\_sense}, \texttt{complex\_instruction},
\texttt{visual\_appearance}, and \texttt{long\_horizon}, with $N=60$
episodes per split ($N=300$ total). The navigation policy is fixed; no
retraining or policy modification is performed. All uncertainty methods
are evaluated post hoc on the same trajectories \cite{yang2025embodiedbench}.

We report standard navigation performance (Success Rate and mean SPL)
together with SRCN ranking metrics:
$\mathrm{AURC}_{\mathrm{SR}}$, $\mathrm{E\text{-}AURC}_{\mathrm{SR}}$ for
success-based loss, and
$\mathrm{AURC}_{\mathrm{SPL}}$, $\mathrm{E\text{-}AURC}_{\mathrm{SPL}}$
for SPL-based loss. Lower values indicate a better ranking of difficult
episodes, with $\mathrm{E\text{-}AURC}=0$ corresponding to oracle
ordering.

Table~\ref{tab:nav_perf_combined} presents baseline navigation performance across three LLM backbones: GPT-4o, GPT-5-mini, and Gemini-1.5-Flash. For GPT-4o, success rates exceed 53\% on four splits, but fall sharply to 16.7\% on \texttt{long\_horizon}, where lengthy execution chains increase compounding errors. Using GPT-5-mini, success exceeds 65\% on four splits, but the success rate for \texttt{log\_horizon} does not improve. The resulting degradation in both Success Rate and
SPL makes this split a stringent test of trajectory-level uncertainty
ranking.

\subsection{SRCN Summary Metrics}

Table~\ref{tab:srcn_sr} reports success-based SRCN results per split and as weighted averages across the five splits ($N=60$ each). 
Trajectory-consistent uncertainty achieves the lowest or tied-lowest $\mathrm{AURC}_{\mathrm{SR}}$ on every split and the best weighted-average performance overall. 
Its excess risk is $\mathrm{E\text{-}AURC}_{\mathrm{SR}}=0.00$ on four splits and $0.01$ on \texttt{long\_horizon}, indicating a near-optimal ordering of failure episodes for all three models: GPT-4o, GPT-5-mini, and Gemini-1.5-Flash.
In contrast, entropy-, conformal-, and self-consistency-based baselines exhibit substantially higher selective risk, reflecting weaker separation between successful and failing trajectories.

The advantage becomes most pronounced on \texttt{long\_horizon} (SR $=16.7\%$ for GPT-4o and GPT-5-mini), where longer execution chains amplify compounding errors. Under these conditions, entropy and conformal signals approach near-random ordering, while trajectory-aware signals remain close to oracle. This indicates that geometric progress dynamics, rather than instantaneous action dispersion, provide the dominant signal for detecting navigation failure under deeper temporal horizons.

Table~\ref{tab:srcn_spl_combined} reports selective risk evaluation based on SPL. Across most splits and models, GroundControl achieves the lowest or near-lowest values, demonstrating its ability to accurately identify trajectories that lead to inefficient navigation paths. For GPT-4o, GroundControl consistently attains the best performance across all splits, achieving the lowest average $\mathrm{AURC}_{\mathrm{SPL}}$ (0.29) and $\mathrm{E\text{-}AURC}_{\mathrm{SPL}}$ (0.01). Similar trends are observed for GPT-5-mini and Gemini-1.5-Flash, where GroundControl again achieves the best average scores, indicating strong cross-model generalization. Although entropy occasionally performs competitively for some GPT-5-mini splits, its performance deteriorates on more challenging settings such as long-horizon tasks. In contrast, conformal, self-consistent, invalid-action, and random baselines exhibit significantly higher AURC and E-AURC values, suggesting a weaker ability to rank trajectories by navigation efficiency. 

\begin{figure}[t!]
  \centering
  \includegraphics[width=\linewidth]{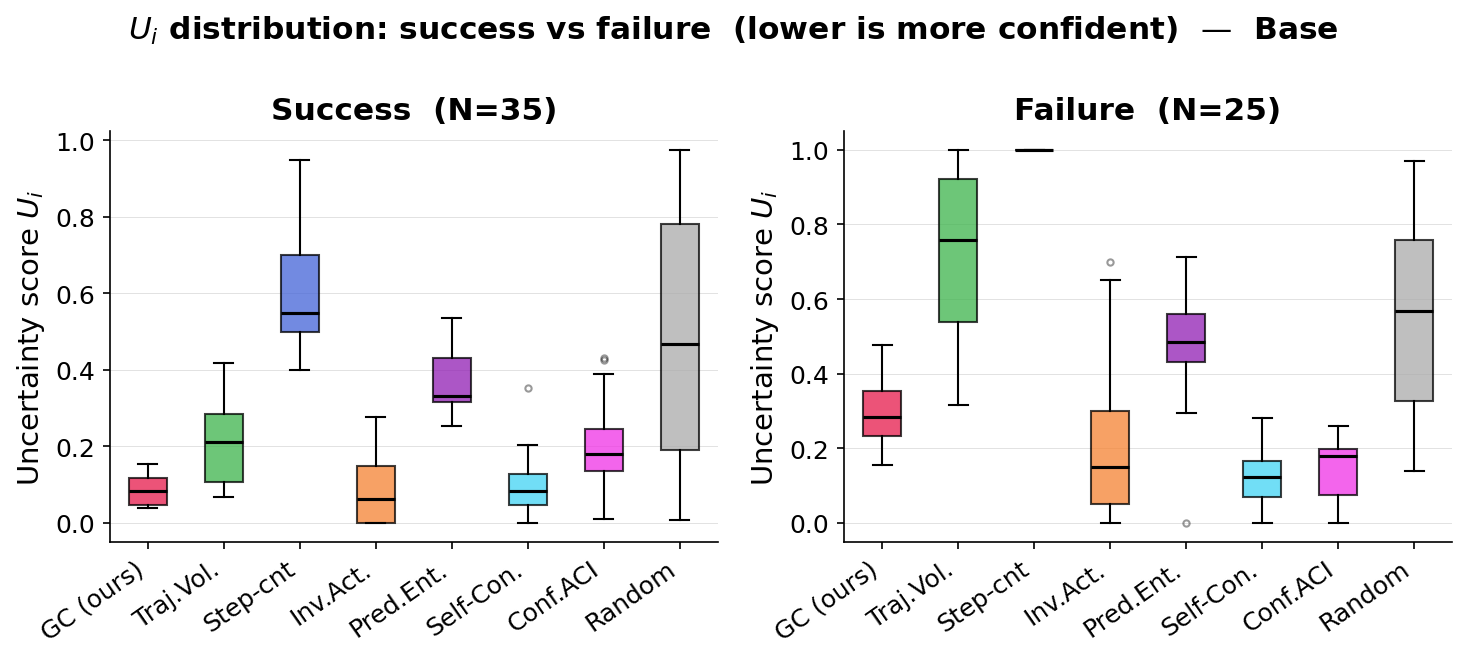}
  \includegraphics[width=\linewidth]{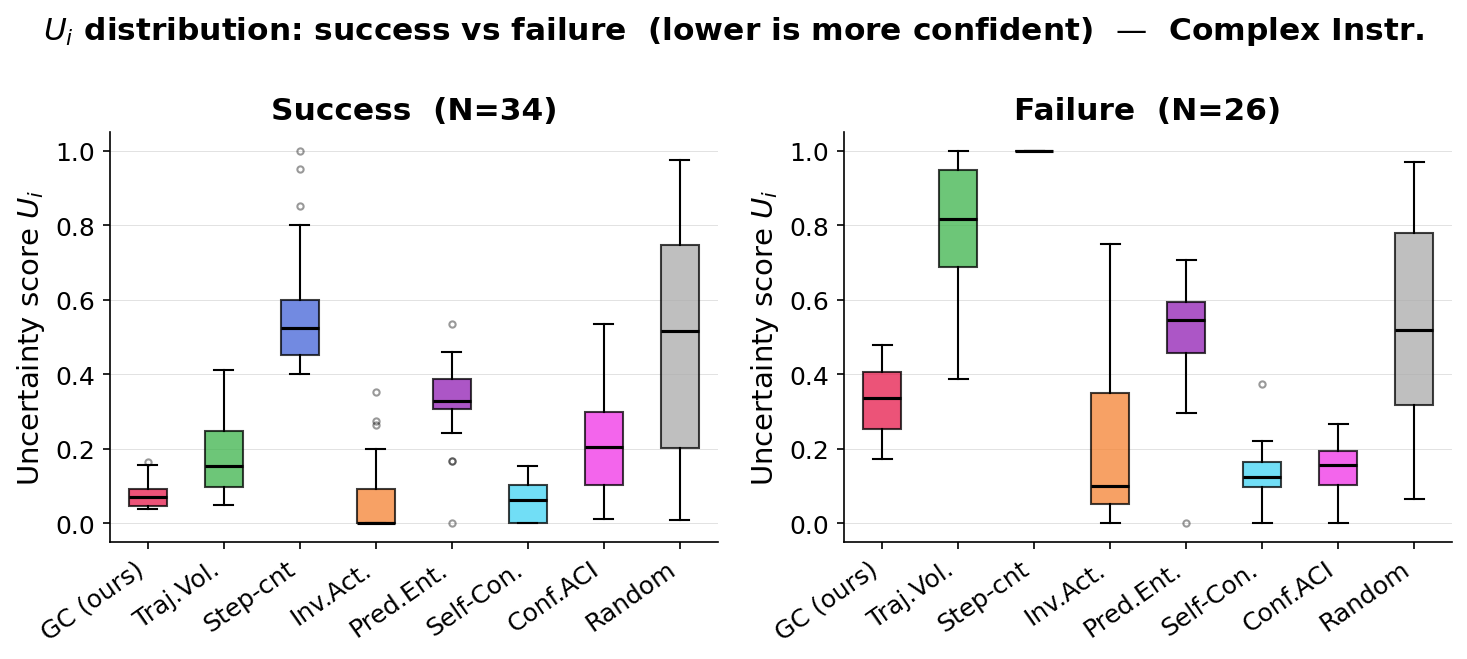}
  \includegraphics[width=\linewidth]{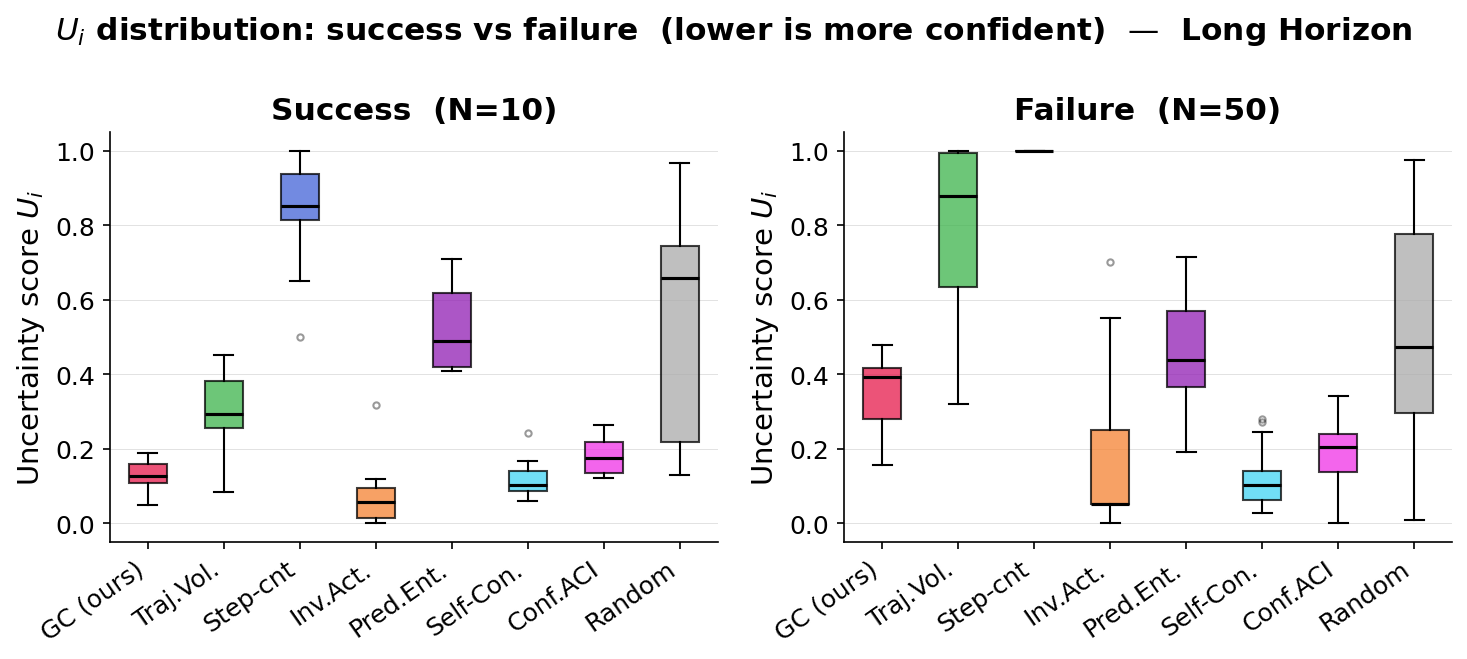}
  \caption{Distribution of episode-level uncertainty $U_i$ conditioned on outcome (left: successes; right: failures) for representative splits (base, complex instruction, and long horizon). A well-ranked estimator shifts failure episodes toward higher $U_i$ with limited overlap. GroundControl consistently produces a clear upward shift and tighter success cluster, particularly under long-horizon compounding effects, whereas entropy- and conformal-based baselines exhibit substantial overlap. This separation aligns with the near-oracle selective ranking observed under SRCN.\vspace{-15pt}}
  \label{fig:box_base}
\end{figure}

\begin{figure*}[t]
  \centering
  \includegraphics[width=0.24\linewidth]{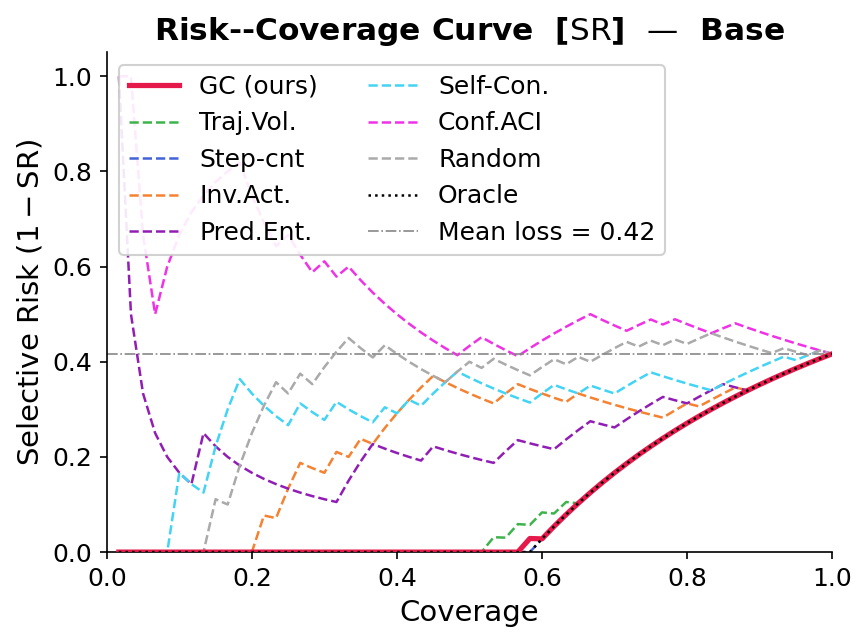}
  \includegraphics[width=0.24\linewidth]{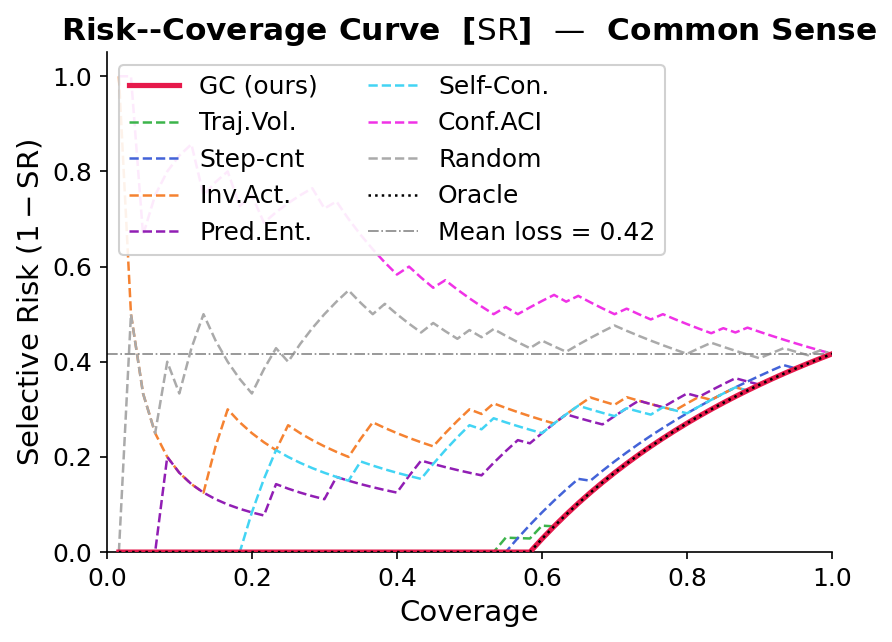}
  \includegraphics[width=0.24\linewidth]{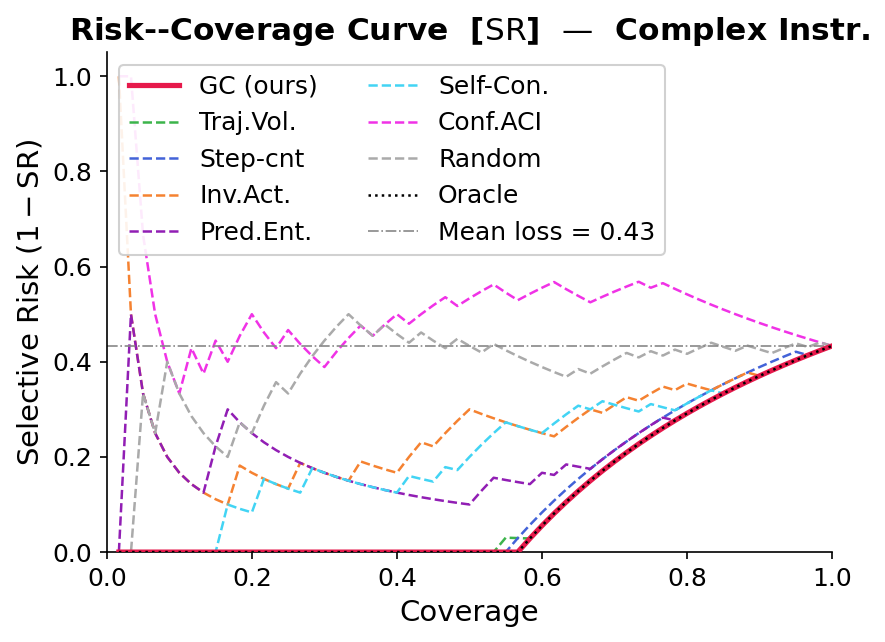}
  \includegraphics[width=0.24\linewidth]{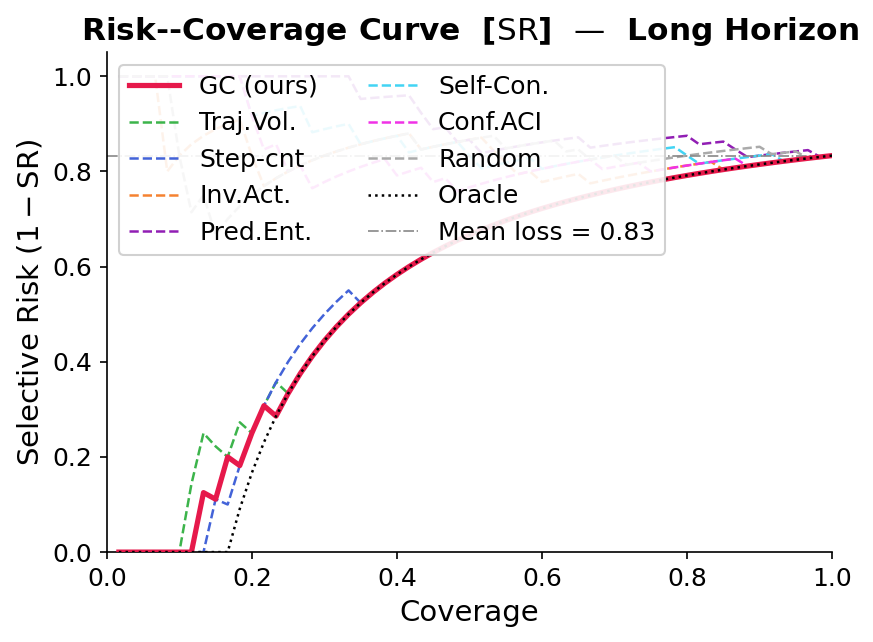}
 \caption{Risk--coverage curve ($N=60$). Episodes are sorted by ascending uncertainty $U_i$ and progressively revealed; the $y$-axis shows selective risk $\hat{R}(\theta)=\frac{1}{|\mathcal{C}(\theta)|}\sum_{i \in \mathcal{C}(\theta)}(1-\mathrm{Succ}_i)$ over the covered subset. The dotted black line denotes the oracle ordering and the dash-dot line the unconditional failure rate. Lower curves indicate better selective ranking. GroundControl closely tracks the oracle.}
  \label{fig:rc_sr_base}
\end{figure*}

\begin{figure*}[t]
  \centering
  \includegraphics[width=0.24\linewidth]{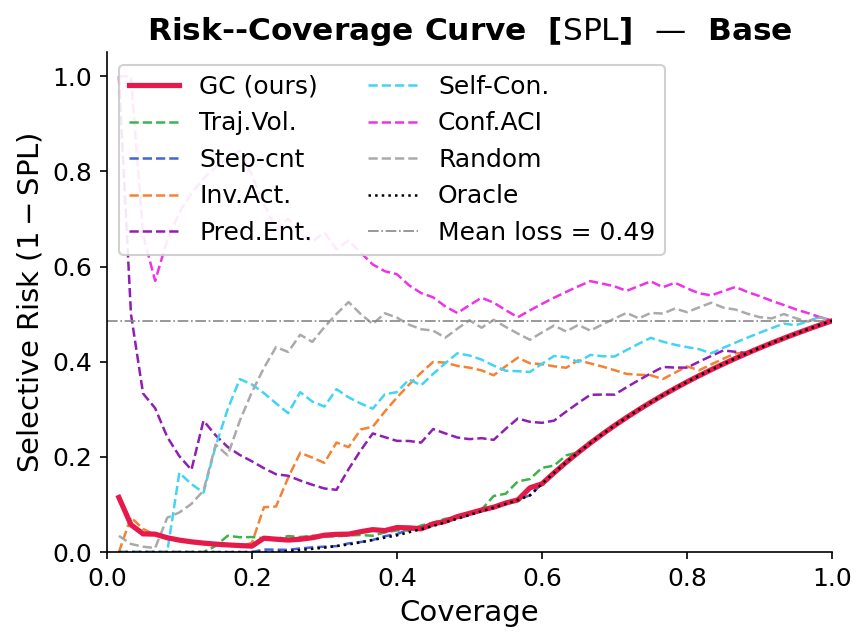}
  \includegraphics[width=0.24\linewidth]{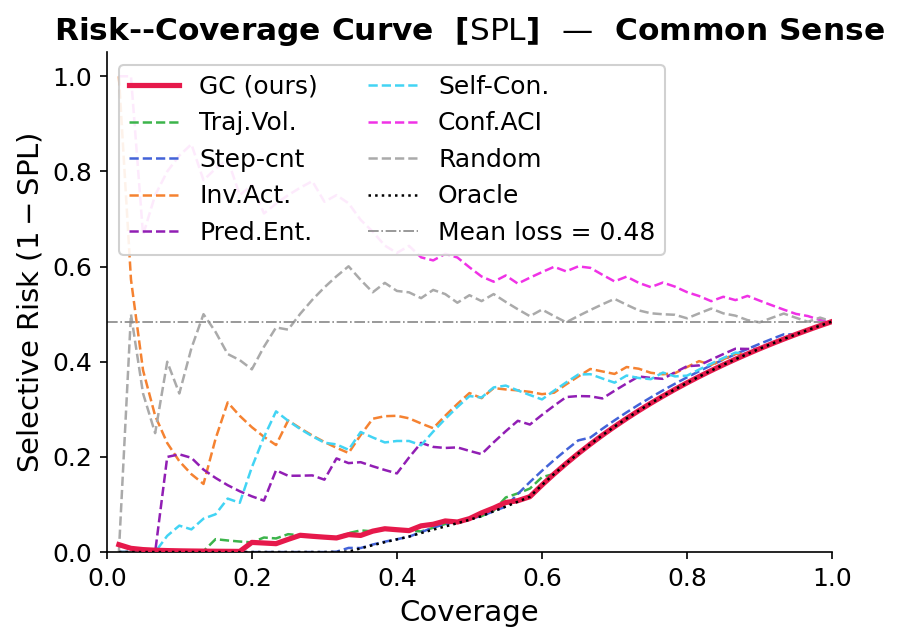}
  \includegraphics[width=0.24\linewidth]{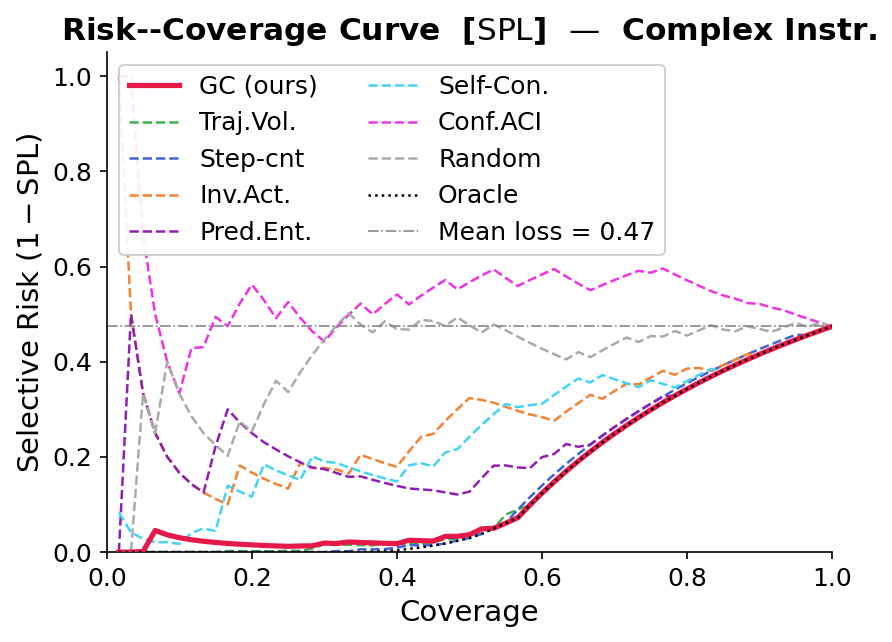}
  \includegraphics[width=0.24\linewidth]{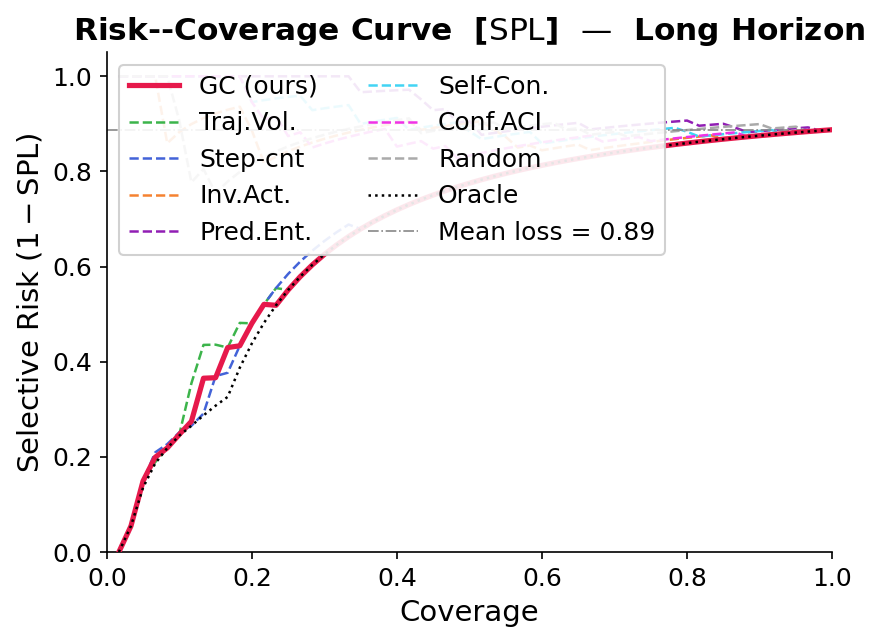}
  \caption{Risk--coverage curve using path-efficiency loss $(1-\mathrm{SPL}_i)$. Unlike binary success risk, SPL-risk captures navigation
  inefficiency: a covered episode contributes a graded penalty proportional to how
  far the agent's path deviates from the shortest route. GroundControl's
  trajectory-level signals, particularly the Kalman innovation energy and
  path-efficiency sensitivity, allow it to maintain a low SPL-risk curve at
  every coverage level.}
  \label{fig:rc_spl_base}
\end{figure*}


\begin{figure*}[t]
  \centering
  \includegraphics[width=\linewidth]{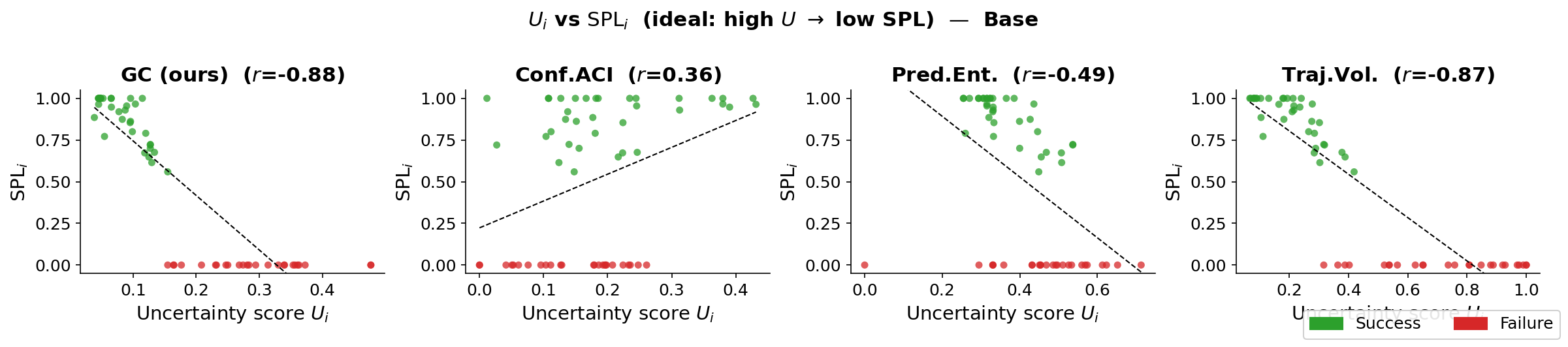}
  \includegraphics[width=\linewidth]{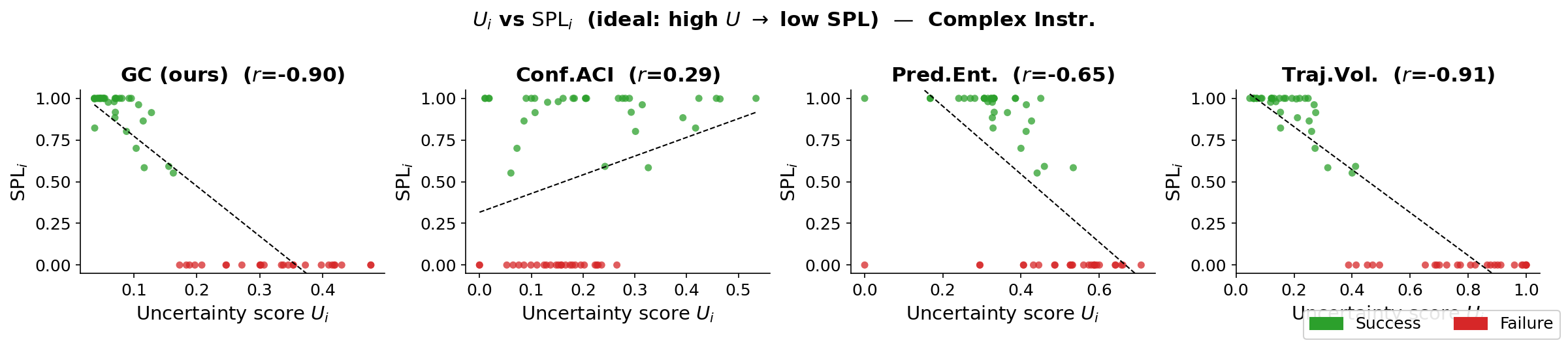}
  \includegraphics[width=\linewidth]{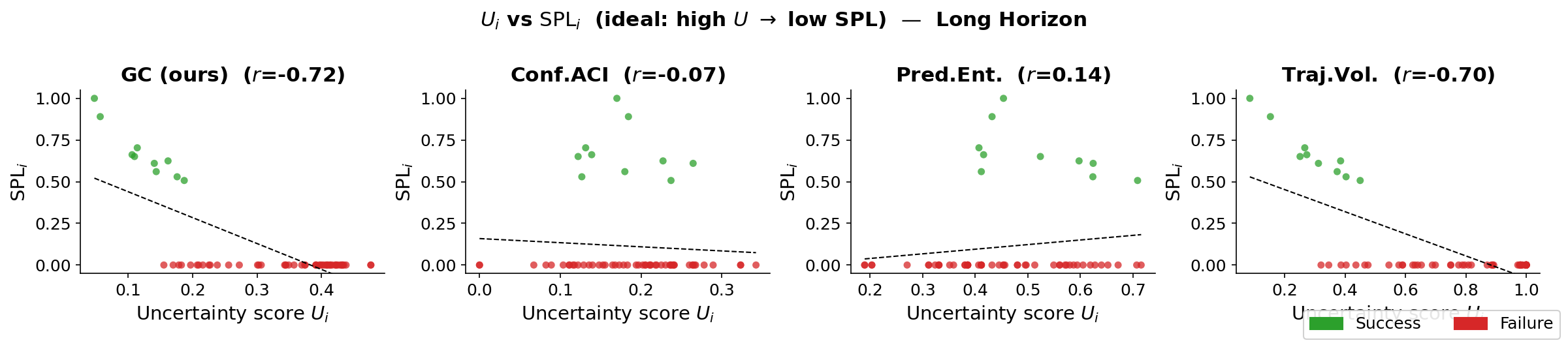}
  \caption{Per-episode uncertainty score $U_i$ versus $\mathrm{SPL}_i$ for each
  method. Points are coloured green (success) and red
  (failure). A well-calibrated estimator produces a negative correlation, high
  $U_i$ for failed/inefficient episodes and low $U_i$ for successful ones shown
  by the dashed linear trend. The Pearson correlation $r$ is annotated in each
  panel. GroundControl achieves the strongest negative correlation among methods
  on this set, indicating that its multi-signal fusion captures both categorical
  (success/failure) and continuous (efficiency) aspects of navigation quality.}
  \label{fig:scatter_base}
\end{figure*}

\subsection{Risk--Coverage Curves and Diagnostic Plots}

Figure~\ref{fig:box_base} shows the distribution of episode-level uncertainty $U_i$ conditioned on outcome for representative splits. 
A reliable estimator should assign consistently higher scores to failing trajectories while maintaining a compact distribution for successful executions. 
The trajectory-consistent formulation produces a clear upward shift of the failure distribution with limited overlap across \texttt{base}, \texttt{complex\_instruction}, and \texttt{long\_horizon}. 
In contrast, entropy- and conformal-based baselines exhibit substantial overlap between success and failure distributions, indicating weak discriminative structure. 
The separation is most pronounced on \texttt{long\_horizon}, where compounding execution errors create sustained deviations in distance dynamics that are captured by the proposed trajectory signal. 
This separation implies that degrading executions can be detected directly from trajectory evolution, enabling runtime intervention before episode termination.

Distributional separation alone does not guarantee correct ranking. 
The risk--coverage curves in Fig.~\ref{fig:rc_sr_base} therefore evaluate ordering quality explicitly. 
Episodes are sorted by ascending $U_i$ and progressively revealed as coverage increases. 
The proposed method closely tracks the oracle boundary under success-based selective risk, maintaining low failure rates even at high coverage. 
This indicates that failing trajectories are consistently assigned higher uncertainty and removed early when a system abstains from high-risk episodes. 
Entropy- and conformal-based baselines exhibit elevated risk at intermediate coverage, reflecting weaker isolation of failing executions.

Figure~\ref{fig:rc_spl_base} shows risk--coverage curves under SPL-based loss, which penalizes inefficient trajectories in addition to outright failures. 
Trajectory-consistent uncertainty maintains low selective risk across coverage levels, indicating sensitivity to gradual degradation in navigation efficiency rather than only terminal failure. 
This behavior is particularly relevant for robotic navigation, where inefficient wandering, oscillatory motion, or repeated backtracking often precede failure and consume limited execution time or energy.

Finally, Fig.~\ref{fig:scatter_base} plots uncertainty $U_i$ against episode efficiency $\mathrm{SPL}_i$. 
The proposed estimator exhibits a strong negative correlation across splits: low-efficiency or failing episodes map to higher uncertainty, while efficient trajectories concentrate at low $U_i$. 
Entropy- and conformal-based baselines show substantially weaker monotonic structure, suggesting that action-level dispersion alone does not reflect trajectory-level execution quality. 
Overall, the alignment between uncertainty and geometric progress supports interpretable runtime introspection capable of identifying degrading navigation behavior before catastrophic failure.

\section{CONCLUSION}

We framed uncertainty in embodied navigation as a trajectory-level consistency problem and introduced Selective Risk--Coverage Navigation (SRCN), a protocol that evaluates how well uncertainty scores rank episodes by failure or inefficiency through risk--coverage analysis. We presented GroundControl, a trajectory-aware estimator based on distance-to-goal dynamics, where a constant-velocity Kalman filter models nominal goal-directed motion and statistically significant deviations are captured through normalized innovation statistics and belief dispersion, combined with global trajectory features. Across EB-Navigation splits, GroundControl achieves near-oracle ordering under success-based selective risk and remains competitive under SPL-based evaluation despite fixed-horizon length artifacts. These results indicate that trajectory-consistent signals capture execution degradation more reliably than entropy or conformal residuals, supporting interpretable introspection for embodied navigation systems.

\bibliographystyle{IEEEtran}
\bibliography{main}

\end{document}